\documentstyle[art11,twocolumn]{article}
\newcommand{\dcg}{\mbox{\sc dcg}}
\newcommand{\hpsg}{\mbox{\sc hpsg}}
\newcommand{\infrule}[3]
           {\frac{\begin{array}{c} #1 \\ #2 \end{array}}
                 {#3}
           }
\newcommand{\lra}[1]  %left and right angle around an mbox
           {\mbox{$ \left\langle \mbox{#1} \right\rangle $}
           }

\title{\sf Bottom-Up Earley Deduction}
\author{Gregor Erbach\thanks{This work was supported by the Deutsche
                             Forschungsgemeinschaft through the
                             project N3 ``Bidirektionale Linguistische
                             Deduktion (BiLD)'' in the
                             Sonderforschungsbereich
                             314 {\it K\"unstliche Intelligenz ---
                             Wissensbasierte Systeme} and by the
                             Commission of the European Communities
                             through the project LRE-61-061 ``Reusable
                             Grammatical Resources.''
                             I would like to thank G\"unter Neumann,
                             Christer Samuelsson and Mats Wir\'en for
                             comments on this paper.
                             }   \\
        University of the Saarland \\
        Computational Linguistics \\
        D-66041 Saarbr\"ucken, Germany \\
        erbach@coli.uni-sb.de
        }
\date{CMP-LG e-print archive cmp-lg/9502004}

\begin{document}

\maketitle

\begin{abstract}
\end{abstract}
We propose a bottom-up variant of Earley deduction. Bottom-up deduction
is preferable to top-down deduction because it allows incremental
processing (even for head-driven grammars), it is data-driven, no
subsumption check is needed, and preference values attached to lexical
items can be used to guide best-first search. We discuss the scanning
step for bottom-up Earley deduction and indexing schemes that
help avoid useless deduction steps.

\section{\rm\hspace{-4mm} Introduction}

Recently, there has been a lot of interest in Earley deduction
\cite{Per:War:83}
with applications to parsing and generation
\cite{Shieber:88,Gerdemann:91,Johnson:93,Doerre:93b}.

Earley deduction is a very attractive framwork for natural
language processing because it has the following properties
and applications.

\begin{itemize}
\item Memoization and reuse of partial results
\item Incremental processing by addition of new items
\item Hypothetical reasoning by keeping track of dependencies
      between items
\item Best-first search by means of an agenda
\end{itemize}

Like Earley's algorithm, all of these approaches operate
top-down (backward chaining).
The interest has naturally focussed on top-down methods because they
are at least to a certain degree goal-directed.

In this paper, we present a bottom-up variant of
Earley deduction, which we find advantageous for the following
reasons:

\begin{description}
\item[Incrementality:] Portions of an input string can be analysed
      as soon as they are produced (or generated as soon as the
      what-to-say component has decided to verbalize them), even
      for grammars where one cannot assume that the left-corner
      has been predicted before it is scanned.
\item[Data-Driven Processing:] Top-down algorithms are not well suited for
      processing grammatical theories like Categorial
      Grammar or \hpsg\ that would only allow very general predictions
      because they
      make use of general
      schemata instead of construction-specific rules.
      For these grammars data-driven
      bottom-up processing is more appropriate. The same is true for
      large-coverage rule-based grammars which lead to the creation of
      very many predictions.
\item[Subsumption Checking:]Since the bottom-up algorithm does not
      have a prediction step, there is no need for the costly operation
      of subsumption checking.%
      \footnote{Subsumption checking may still be needed to filter
                out spurious ambiguities.}
\item[Search Strategy:] In the case where lexical entries have been
      associated with preference information,
      this information
      can be exploited to guide the heuristic search.
\end{description}

\section{\rm \hspace{-4mm} Bottom-up Earley Deduction}

Earley deduction \cite{Per:War:83} is based on grammars encoded
as definite clauses.
The instantiation (prediction) rule of top-down Earley deduction is
not needed in bottom-up Earley deduction, because there is no prediction.
There is only one inference rule, namely the
reduction rule (\ref{reduction-rule}).%
\footnote{This rule is called {\bf combine} by Earley, and is also
          referred to as the {\bf fundamental rule} in the literature
          on chart parsing.}
In (\ref{reduction-rule}), $X$, $G$ and $G'$ are literals,
$\Omega$ is a (possibly empty)
sequence of literals, and
$\sigma$ is the most general unifier of $G$ and $G'$.
The leftmost literal in the body of a non-unit clause is always
the selected literal.

\begin{equation} \label{reduction-rule}
\infrule{X \leftarrow G \wedge \Omega}
        {G' \leftarrow}
        {\sigma(X \leftarrow \Omega)}\\
\end{equation}

In principle, this rule can be applied to any pair of unit clauses
and non-unit clauses of the program to derive any consequences of
the program. In order to reduce this search space and achieve a more
goal-directed behaviour, the rule is not applied to any pair of
clauses, but clauses are only selected if they can contribute to
a proof of the goal. The set of selected clauses is called the
{\em chart.\/}%
\footnote{The chart differs from the {\em state} of \cite{Per:War:83}
          in that clauses in the chart are indexed (cf.
          section \ref{idx}).}
The selection of clauses is guided by a scanning
step (section \ref{scan}) and indexing of clauses (section \ref{idx}).

\subsection{\rm Scanning} \label{scan}

The purpose of the scanning step, which corresponds to lexical
lookup in chart parsers, is to look up base cases of recursive
definitions to serve as a starting point for bottom-up processing.
The scanning step selects clauses that can
appear as leaves in the proof tree for a given goal $G$.

Consider the following simple definition of an {\sc hpsg},
with the recursive definition of the predicate {\tt sign/1}.%
\footnote{We use feature terms in definite clauses in addition to
          Prolog terms. {\tt f:X}
          denotes a feature structure where {\tt X} is the value
          of feature {\tt f},
          and {\tt X \& Y} denotes the conjunction of the feature terms
          {\tt X} and {\tt Y}.}

\begin{small}
\begin{verbatim}
sign(X) <- phrasal_sign(X).
sign(X) <- lexical_sign(X).

phrasal_sign(X & dtrs:(head_dtr:HD &
                       comp_dtr:CD)  ) <-
     sign(HD),
     sign(CD),
     principles(X,HD,CD).

principles(X,HD,CD) <-
     constituent_order_principle(X,HD,CD),
     head_feature_principle(X,HD),
     ...

constituent_order_principle(phon:X_Ph,
                            phon:HD_Ph,
                            phon:CD_Ph) <-
     sequence_union(CD_Ph,HD_Ph,X_Ph).
\end{verbatim}
\end{small}

The predicate {\tt sign/1} is defined recursively, and the base case
is the predicate {\tt lexical\_sign/1}.
But, clearly it is not restrictive enough to find only the predicate
name of the base case for a given goal. The base cases must also be
instantiated in order to find those that are useful for proving a given
goal. In the case of parsing, the lookup of base cases
(lexical items) will depend on
the words that are present in the input string. This is implied by
the first goal of the predicate {\tt principles/3}, the
{\tt constituent order principle}, which determines how the {\sc phon}
value of a constituent is constructed from the {\sc phon} values of
its daughters. In general, we assume that the
constituent order principle makes use of a linear and non-erasing
operation for combining strings.%
\footnote{There is an obvious connection to the Linear Context-Free
          Rewriting Systems (LCFRS) \cite{Vij:Wei:Jos:87,Weir:88}.}
If this is the case, then all the words contained in the
{\sc phon} value of the goal can have their lexical items selected
as unit clauses to start bottom-up processing.

For generation, an analogous condition on logical forms has been
proposed by Shieber \cite{Shieber:88} as the ``semantic monotonicity
condition,'' which requires that the logical form of every base case must
subsume some portion of the goal's logical form.

Base case lookup must be defined specifically for different
grammatical theories and directions of processing
by the predicate {\tt lookup/2}, whose first argument is the
goal and whose second argument is the selected base case.
The following clause defines the {\tt lookup} relation for parsing with \hpsg .

\begin{small}
\begin{verbatim}
% lookup(+Goal,-BaseCase)
lookup(sign(phon:PhonList),
       lexical_sign(phon:[Word] & synsem:X)
      ) <-
     member(Word,PhonList),
     lexicon(Word,X).
\end{verbatim}
\end{small}

Note that the base case clauses can become further instantiated in
this step. If concatenation (of difference lists) is used as the
operation on strings, then each base case clause can be instantiated
with the string that follows it.
This avoids combination of items that are not
adjacent in the input string.

\begin{small}
\begin{verbatim}
lookup(sign(phon:PhonList),
       lexical_sign(phon:[Word|Suf]-Suf &
                    synsem: Synsem)
      ) <-
     append(_,[Word|Suf],PhonList),
     lexicon(Word,Synsem).
\end{verbatim}
\end{small}

In bottom-up Earley deduction, the first step towards proving a goal
is perform lookup for the goal, and to add all the resulting (unit)
clauses to the chart.
Also, all non-unit clauses of the program, which
can appear as internal nodes in the proof tree of the goal, are
added to the chart.

The scanning step achieves a certain degree of goal-directedness
for bottom-up algorithms because only those clauses which can appear as
leaves in the proof tree of the goal are added to the chart.

\subsection{\rm Indexing} \label{idx}

An item in normal context-free chart parsing can be regarded as a pair
\lra{R,S} consisting of a dotted rule R and the substring S that the
item covers (a pair of starting and ending position). The fundamental
rule of chart parsing makes use of these string positions to ensure
that only adjacent substrings are combined and that the result is
the concatenation of the substrings.

In grammar formalisms like \dcg\ or \hpsg , the complex nonterminals
have an argument or a feature {(\sc phon)} that represents
the covered substring explicitly. The combination of the substrings
is explicit in the rules of the grammar.
As a consequence, Earley deduction does
not need to make use of string positions for its clauses, as Pereira and Warren
\cite{Per:War:83} point out.

Moreover, the use of string positions known from chart parsing is
too inflexible because it allows only concatenation of adjacent
contiguous substrings. In linguistic theory, the interest has shifted from
phrase structure rules that combine adjacent and contiguous constituents to

\begin{itemize}
   \item principle-based approaches to grammar that state
         general well-formedness conditions instead of describing
         particular constructions (e.g. \hpsg )
   \item operations on strings that go beyond concatenation
         (head wrapping \cite{Pollard:84}, tree adjoining
\cite{Vij:Wei:Jos:87},
         sequence union \cite{Reape:90}).
\end{itemize}

The string positions known from chart parsing are also inadequate
for generation, as pointed out by Shieber \cite{Shieber:88} in whose
generator all items go from position 0 to 0 so that any item can
be combined with any item.

However, the string positions are useful as an indexing of the items
so that it can be easily detected whether their combination can
contribute to a proof of the goal. This is especially important for a
bottom-up algorithm which is not goal-directed like top-down
processing. Without indexing, there are too many combinations of items
which are useless for a proof of the goal, in fact there may be infinitely
many items so that termination problems can arise.

For example, in an order-monotonic grammar formalism
that uses sequence union as the
operation for combining strings, a combination of items would be
useless which results in a sign in which the words are not in
the same order as in the input string \cite{Noord:93}.

We generalize the indexing scheme from chart parsing
in order to allow different operations for the combination of strings.
Indexing improves efficiency by detecting combinations that would fail
anyway and by avoiding combinations of items that are useless for
a proof of the goal.

We define an item as a pair of a clause $Cl$ and an index $Idx$,
written as
$ \left\langle \mbox{\em Cl, Idx} \right\rangle $.

Below, we give some examples of possible indexing schemes.
Other indexing schemes can be used if they are needed.

\begin{description}
\item [1. Non-reuse of Items:] This is useful for LCFRS, where no word
       of the input string can be used twice in a proof, or for generation
       where no part of the goal logical form should be verbalized twice
       in a derivation.
\item [2. Non-adjacent combination:] This indexing scheme is useful for
       order-monotonic grammars.
\item [3. Non-directional adjacent combination:] This indexing is used if
       only adjacent constituents can be combined, but the order of
       combination is not prescribed (e.g. non-directional basic
       categorial grammars).
\item [4. Directional adjacent combination:] This is used for grammars
       with a ``context-free backbone.''
\item [5. Free combination:] Allows an item to be used several times in
       a proof, for example for the non-unit clauses of the program, which
       would be represented as items of the form
       $ \left\langle X \leftarrow
         G_1 \wedge \ldots \wedge G_n, \mbox{free} \right\rangle $.
\end{description}

The following table summarizes the properties of these five
combination schemes. {\em Index 1 (I1)} is the index associated with
the non-unit clause, {\em Index 2 (I2)} is associated with the unit
clause, and $I1 \star I2$ is the result of combining the indices.

\vspace{3mm}

\noindent
\begin{small}
\begin{tabular}{|l|c|c|c|l|} \hline
   & {\em Index 1} & {\em Index 2} &
     {\em Result} & {\em Note}                             \\
   & $I1$  & $I2$   & $I1 \star I2$ &                      \\ \hline
1. & $X$   & $Y$    & $X \cup Y$  & $X \cap Y = \emptyset$ \\ \hline
2. & $X$   & $Y$    & $X \odot Y$ &                        \\ \hline
3. & $X+Y$ & $Y+Z$  & $X+Z$       &                        \\
   & $Y+Z$ & $X+Y$  & $X+Z$       &                        \\ \hline
4. & $X-Y$ & $Y-Z$  & $X-Z$       &                        \\ \hline
5. & $X$   & `free' & $X$         &                        \\
   & `free' & $X$   & $X$         &                        \\ \hline
\end{tabular}
\end{small}

\vspace{3mm}

In case 2 (``non-adjacent combination''), the indices $X$ and
$Y$ consist of a set of string positions, and the operation $\odot$
is the union of these string positions, provided that no two string
positions from $X$ and $Y$ do overlap.

In (\ref{item-rule}), the reduction rule is
augmented to handle indices. $X \star Y$ denotes
the combination of the indices $X$ and $Y$.

\begin{equation} \label{item-rule}
\infrule{\lra{$X \leftarrow G \wedge \Omega, I1$}}
        {\lra{$G' \leftarrow, I2$}}
        {\lra{$\sigma(X \leftarrow \Omega),I1 \star I2$}}\\
\end{equation}

With the use of indices, the {\tt lookup} relation becomes a
relation between goals and {\em items.} The following
specification of the {\tt lookup} relation provides indexing according
to string positions as in a chart parser (usable for combination
schemes 2, 3, and 4).

\begin{small}
\begin{verbatim}
lookup(sign(phon:PhonList),
       item(lexical_sign(phon:[Word] &
                         synsem:X),
            Begin-End)
      ) <-
     nth_member(Word,Begin,End,PhonList),
     lexicon(Word,X).

nth_member(X,0,1,[X|_]).
nth_member(X,N1,N2,[_|R]) <-
     nth_member(X,N0,N1,R),
     N2 is N1 + 1.
\end{verbatim}
\end{small}

\subsection{\rm Goal Types}

In constraint-based grammars there are some predicates that are
not adequately dealt with by bottom-up Earley deduction, for example
the Head Feature Principle and the Subcategorization
Principle of \hpsg . The Head Feature Principle just unifies two
variables, so that it can be executed at compile time
and need not be called as a goal at runtime. The Subcategorization
Principle involves an operation on lists ({\tt append/3 \rm or
\tt delete/3} in different
formalizations)
that does not need
bottom-up processing, but can better be evaluated by top-down resolution
if its arguments are sufficiently instantiated. Creating
and managing items for these proofs is too much of a computational
overhead, and, moreover, a proof may not terminate in the bottom-up case
because
infinitely many consequences may be derived from the base case of
a recursively defined relation.

In order to deal with such goals, we associate the goals in the body
of a clause with goal types. The goals that are relevant for bottom-up
Earley deduction are called {\em waiting goals} because they wait
until they are activated by a unit clause that unifies with the
goal.%
\footnote{The other goal types are top-down goals (top-down depth-first
          search), $x$-corner goals (which combine bottom-up and top-down
          processing like left-corner or head-corner algorithms), Prolog
          goals (which are directly executed by Prolog for efficiency or
          side-effects), and chart goals which create a new, independent
          chart for the proof of the goal.
          D\"orre \cite{Doerre:93b} proposes a system with two goal types,
          namely {\em trigger goals}%
          %(which correspond to our waiting goals)
          , which lead to the creation of items
          and other goals which don't.}
Whenever a unit clause is combined with a non-unit clause
all goals up to the first waiting goal of the resulting clause are
proved according to their goal type, and then a new clause is
added whose selected goal is the first waiting goal.

In the following
inference rule for clauses with mixed goal types,
$\Xi$ is a (possibly empty) sequence of goals without any waiting goals,
and $\Omega$ is a (possibly empty) sequence of goals starting with a
waiting goal. $\sigma$ is
the most general unifier of $G$ and $G'$,
and the substitution $\tau$ is the solution
which results from proving the sequence of goals $\Xi$.

\begin{equation} \label{mixed-rule}
\infrule{\lra{$X \leftarrow G \wedge \Xi \wedge \Omega,I1$}}
        {\lra{$G' \leftarrow,I2$}}
        {\lra{$\tau\sigma (X \leftarrow \Omega),I1 \star I2$}}\\
\end{equation}

\subsection{\rm Correctness and Completeness}

In order to show the correctness of the system, we must show
that the scanning step only adds consequences of the program
to the chart, and that any items derived by the inference rule
are consequences of the program clauses. The former is easy to
show because all clauses added by the scanning step are {\em instances}
of program clauses, and the inference rule performs a resolution
step whose correctness is well-known
in logic programming. The other goal types are also proved
by resolution.

There are two potential sources of incompleteness in the algorithm. One is
that the scanning step may not add all the program clauses to the chart
that are needed for proving a goal, and the other is that
the indexing may prevent the derivation of a clause that is needed
to prove the goal.

In order to avoid incompleteness, the scanning step must add {\em all}
program clauses that are needed for a proof of the goal to the
chart, and the combination of indices may only fail for inference
steps which are useless for a proof of the goal.
That the lookup relation and the indexing scheme satisfy this
property must be shown for particular grammar formalisms.

In order to keep the search space small (and finite to ensure
termination) the scanning
step should (ideally) add {\em only} those items that are needed for
proving the goal to the chart, and the indexing should be chosen
in such a way that it excludes derived items that are useless for
a proof of the goal.

\section{\rm \hspace{-4mm} Best-First Search}

For practical NL applications, it is desirable to have a best-first
search strategy, which follows the most promising paths in the search
space first, and finds preferred solutions before the less preferred ones.

There are often situations where the criteria to guide the search
are available only for the base cases, for example

\begin{itemize}
\item weighted word hypotheses from a speech recognizer
\item readings for ambigous words with probabilities, possibly
      assigned by a stochastic tagger (cf. \cite{Brew:93})
\item hypotheses for correction of string errors which
      should be delayed \cite{Erbach:94} %maybe not
\end{itemize}

Goals and clauses are associated with preference values that are
intended to model the degree of confidence that a particular solution
is the `correct' one. Unit clauses are associated with a numerical
preference value, and non-unit clauses with a formula that determines
how its preference value is computed from the preference values of
the goals in the body of the clause. Preference values can (but need not) be
interpreted as
probabilities.%
\footnote{For further details and examples see \cite{Erbach:93} and
          \cite{Erbach:94}.}

The preference values are the basis for giving priorities to items. For
unit clauses, the priority is identified with the preference value.
For non-unit clauses,
where the preference formula may contain uninstantiated variables, the priority
is the value of the formula with the free variables instantiated to
the highest possible preference value (in case of an interpretation as
probabilities: 1), so that the priority is equal to the maximal
possible preference value for the clause.%
\footnote{There are also other methods for assigning priorities to
          items.} % cite{erbach91} ???

The implementation of best-first search does not combine new items
with the chart immediately, but makes use of an agenda
\cite{kay80}, on which new items are ordered in order of descending
priority. The following is the algorithm for bottom-up
best-first Earley deduction.

%\begin{figure}
\begin{small}
\begin{sf}
\begin{tabbing}
-- \= -- \= -- \= -- \=  \kill
{\bf procedure} prove({\em Goal\/}): \\
-- initialize-agenda({\em Goal\/}) \\
-- consume-agenda          \\
-- for any item \lra{$G$,$I$\/} \\
\> -- return mgu({\em Goal,G\/}) as solution if it exists \\
\\
{\bf procedure} initialize-agenda({\em Goal\/}):    \\
-- for every unit clause {\em UC} in lookup({\em Goal,UC\/}) \\
\> -- create the index $I$ for {\em UC}                \\
\> -- add item $\left\langle UC, I \right\rangle $ to agenda \\
-- for every non-unit program clause $H \leftarrow Body$ \\
\> -- add item \lra{$H \leftarrow Body$,free} to agenda \\
\\
{\sf\bf procedure} add item $I$ to agenda \\
-- compute the priority of $I$ \\
-- agenda $:=$ agenda $\cup \left\{ I \right\} $  \\
\\
{\bf procedure} consume-agenda \\
-- while agenda is not empty \\
\> -- remove item $I$ with  highest priority from agenda \\
\> -- add item $I$ to chart  \\
\\
{\bf procedure} add item \lra{$C,I1$} to chart \\
-- chart := chart  $\cup \left\{ \lra{$C,I1$} \right\} $  \\
-- if $C$ is a unit clause \\
\> -- for all items \lra{$H \leftarrow G \wedge \Xi \wedge \Omega, I2$} \\
\> \> -- if $I = I2 \star I1$ exists \\
\> \> \> and $\sigma = \mbox{mgu}(C,G)$ exists \\
\> \> \> and goals $\Xi$ are provable with solution $\tau$ \\
\> \> \> then add item \lra{$\tau\sigma(H \leftarrow \Omega),I$} to agenda \\
-- if $C =  H \leftarrow G \wedge \Xi \wedge \Omega$ is a non-unit clause \\
\> -- for all items \lra{$G' \leftarrow,\ I2$} \\
\> \> -- if $I = I1 \star I2$ exists \\
\> \> \> and $\sigma = \mbox{mgu}(G,G')$ exists \\
\> \> \> and goals $\Xi$ are provable with solution $\tau$ \\
\> \> \> then add item \lra{$\tau\sigma(H \leftarrow \Omega),I$} to agenda \\

\end{tabbing}
\end{sf}
\end{small}
%\caption{Algorithm for Bottom-Up Best-First Earley Deduction}
%%\label{algorithm}
%\end{figure}

The algorithm is parametrized with respect to
the relation {\tt lookup/2} and the choice of the indexing scheme,
which are specific for different grammatical theories
and directions of processing.

\section{\rm \hspace{-4mm} Implementation}

The bottom-up Earley deduction algorithm described here has been implemented
in Quintus Prolog as part of the GeLD system. GeLD (Generalized Linguistic
Deduction) is an extension of Prolog which provides
typed feature descriptions and preference values as additions to the
expressivity of the language,
and partial evaluation, top-down, head-driven, and bottom-up
Earley deduction as processing strategies.
Tests of the system with small grammars
have shown promising results, and a medium-scale \hpsg\ for German is
presently being implemented in GeLD.
The lookup relation and the choice of an indexing scheme must be
specified by the user of the system.

\section{\rm \hspace{-4mm} Conclusion and Future Work}

We have proposed bottom-up Earley deduction as a useful alternative
to the top-down methods which require
subsumption checking and restriction to avoid prediction loops.

The proposed method should be improved in two directions. The first
is that the lookup predicate should not have to be specified by the user,
but automatically inferred from the program.

The second problem is that all non-unit clauses of the program
are added to the chart. The addition of non-unit clauses should
be made dependent on the goal and the base cases
in order to go from a purely
bottom-up algorithm to a directed algorithm that combines
the advantages of top-down and bottom-up processing.
It has been repeatedly noted \cite{kay80,wiren87,Bou:Noo:93} that
directed methods are more efficient than pure top-down or bottom-up
methods. However, it is not clear how well the
directed methods are applicable to grammars which do not depend on
concatenation and have no unique `left corner' which should be connected
to the start symbol.

It remains to be seen how bottom-up Earley deduction compares with
(and can be combined with) the improved top-down Earley deduction
of D\"orre \cite{Doerre:93b}, Johnson \cite{Johnson:93}
and Neumann \cite{Neummy:94}, and to
head-driven methods with well-formed substring tables \cite{Bou:Noo:93},
and which methods are best suited for which kinds of problems
(e.g. parsing, generation, noisy input, incremental processing etc.).

%\bibliographystyle{plain}
%\bibliography{gor}

\end{document}